\documentclass[10pt,twocolumn,twoside]{IEEEtran}
\usepackage{times}
\usepackage{epsfig}
\usepackage{amsmath}
\usepackage{amssymb}
\usepackage{wasysym}
\usepackage{color}
\usepackage{multirow}
\usepackage{cite}
\usepackage{graphicx} % more modern
\usepackage{subfigure}

\usepackage{algorithm}
\usepackage{algorithmic}

\usepackage{url}
\usepackage[numbers]{natbib}
\DeclareUnicodeCharacter{FB01}{fi}
\usepackage{graphicx} % more modern
\usepackage{subfigure}
\usepackage{color}
\usepackage{lettrine}
\usepackage{mathrsfs}
\usepackage{amsmath}
\usepackage{amsfonts}
\usepackage{xr}
\usepackage{bm}
\usepackage{multirow}

\begin{document}
\title{Temporal Transformer Networks with Self-Supervision for Action Recognition}

\author{
Yongkang~Zhang,~
Jun~Li*,~
Guoming~Wu,~
Han~Zhang,~\\
Zhiping~Shi,~\IEEEmembership{Member,~IEEE,}
Zhaoxun~Liu,~
Zizhang~Wu,~
\thanks{
Y. Zhang, J. Li, G. Wu, H. Zhang, and Z. Shi are with the Information Engineering College, Capital Normal University, Beijing, 100048, China (*Corresponding author: Jun Li, email: junmuzi@gmail.com). 

Z. Liu is with the State Key Lab of Software Development Environment, School of Computer Science and Engineering, Beihang University, Beijing, 100191, China.

Z. Wu is with the Computer Vision Perception Department of ZongMu Technology, Shanghai, 201203, China.
}% <-this % stops a space
}

\markboth{IEEE Transactions on Image Processing,~Vol.~X, No.~X, XXX~2021}%
{Zhang \MakeLowercase{\textit{et al.}}: Temporal Transformer Networks with Self-Supervision for Action Recognition}

\IEEEcompsoctitleabstractindextext{%
\begin{abstract}
In recent years, 2D Convolutional Networks-based video action recognition has encouragingly gained wide popularity; However, constrained by the lack of long-range non-linear temporal relation modeling and reverse motion information modeling, the performance of existing models is, therefore, undercut seriously. To address this urgent problem, we introduce a startling Temporal Transformer Network with Self-supervision (TTSN). Our high-performance TTSN mainly consists of a temporal transformer module and a temporal sequence self-supervision module. Concisely speaking, we utilize the efficient temporal transformer module to model the non-linear temporal dependencies among non-local frames, which significantly enhances complex motion feature representations. The temporal sequence self-supervision module we employ unprecedentedly adopts the streamlined strategy of ``random batch random channel'' to reverse the sequence of video frames, allowing robust extractions of motion information representation from inversed temporal dimensions and improving the generalization capability of the model. Extensive experiments on three widely used datasets (HMDB51, UCF101, and Something-something V1) have conclusively demonstrated that our proposed TTSN is promising as it successfully achieves state-of-the-art performance for action recognition.
\end{abstract}

}

\maketitle

\IEEEdisplaynotcompsoctitleabstractindextext

\IEEEpeerreviewmaketitle

\section{Introduction}
In recent years, deep neural network-based video action recognition methods have undergone considerable progress~\cite{2014Large,simonyan2014two,2016Temporal,8902002,9204452}; notwithstanding, with extra dimensions in videos, it is still challenging to devise a high-performance recognition approach~\cite{2020Self2,li2020spatio}. An essential key to the breakthrough of this issue lies in the understanding of motion information derived from given videos. This ability requires the neural network to possess a solid modeling ability to thoroughly seize the motion information representation in the temporal dimension of a given video. To contrive to meet this requirement, the video action recognition algorithm~\cite{2017Appearance} principally incorporates optical flow-based methods~\cite{2016Temporal}, 3D CNNs-based methods~\cite{2014Learning, 20133D}, and 2D CNNs with temporal modeling methods~\cite{simonyan2014two,2020TDN}. Regarding the optical flow-based methods applied, the temporal sequence of motion information is modeled by optical flows; nevertheless, a lapse of these methods alone is that the optical flow calculation demands numerous overheads, making it not so hospitable to apply. Subsequently, the 3D CNNs-based methods extract motion information directly from RGB frames, end-to-end, by 3D convolution; however, it is unwantedly costly to compute and deploy such a network. Last but not least, the 2D CNNs-based methods, such as RNN~\cite{2015Beyond,2017TORNADO}, model a given video into an ordered sequence of frames; yet they focus solely on capturing crude temporal structure relationship among frames~\cite{li2020spatio}, making the 2D CNNs-based methods alone not so ideal to model motion information effectively and robustly.

\begin{figure}[tp!]
    \centering
    \includegraphics[width=1\linewidth]{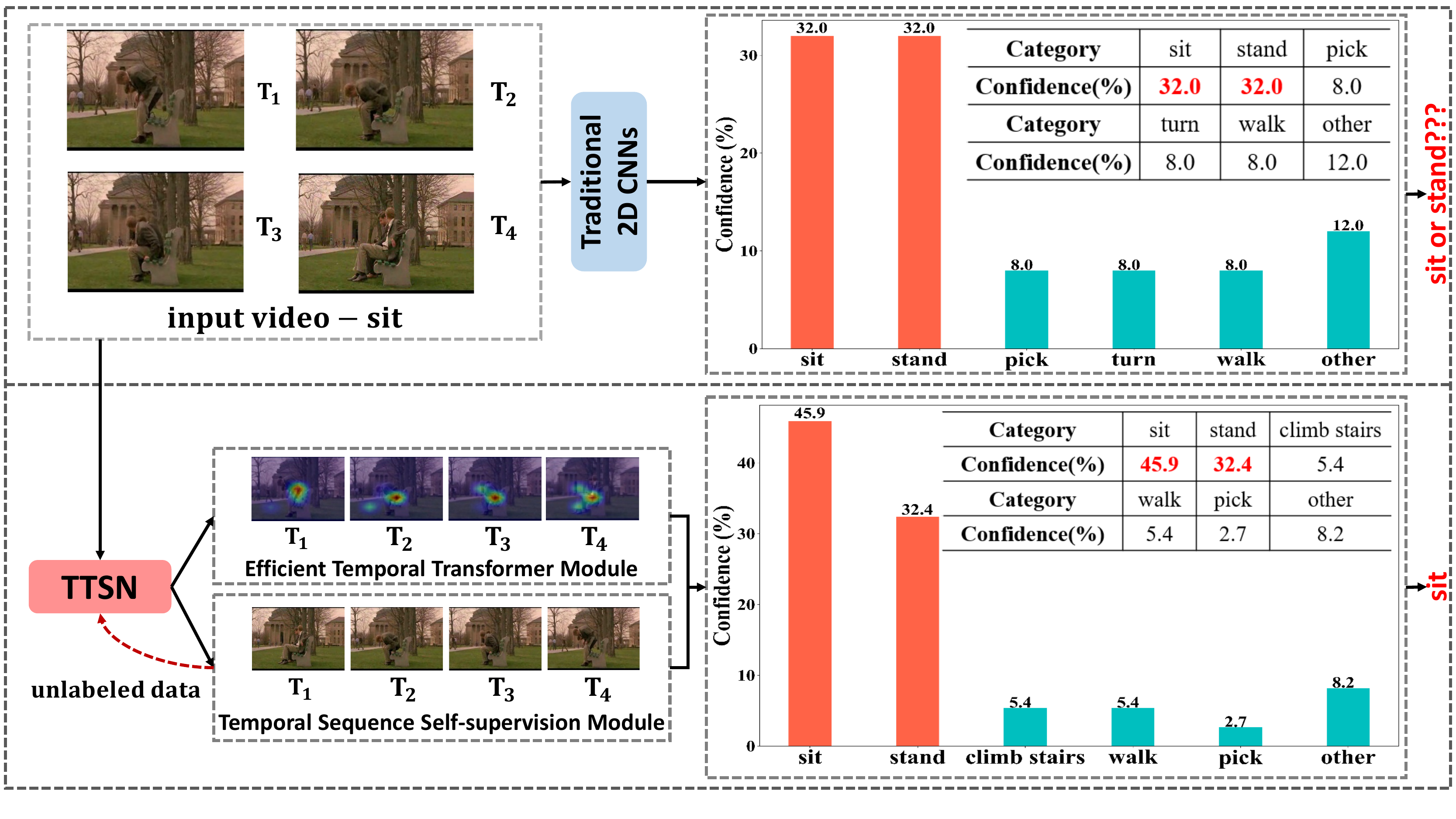}\\
     \caption{An example comparison of recognition results between traditional 2D CNNs (e.g., TDN \cite{2020TDN}) and our TTSN model in the inference stage on the HMDB51 dataset. We notice that the actions (e.g., ``sit'' and ``stand'') that are reverse-ordered in the temporal dimension but similar in the spatial dimension are very confusing to the traditional 2D CNNs. When inputting a video clip with the ``sit'' category, traditional 2D CNNs, lacking effective modeling of the temporal dimension, recognize it as ``sit'' with 32.0\% confidence and ``stand'' with a veritably same-weighted 32.0\% confidence, making the model fail to make the correct prediction. However, our TTSN model, adaptively combining the temporal transformer module and the Temporal Sequence Self-Supervision module and modeling the complex and non-linear temporal motion features over long-range frames, recognizes it as ``sit'' with 45.9\% confidence and ``stand'' with 32.4\% confidence, making the two actions significantly distinguishable.}
    \label{fig:1}
\end{figure}

Delightfully, attempts have been made~\cite{2018TSM,2020TDN,2019STM,2016Temporal,2020TEA} to, via 2D CNNs,  directly extract motion information representations from RGB frames,  having assisted in lifting the above issues concerning recognition accuracy and computation complexity. However, most of these methods model temporal dimensions base singularly on first-order temporal differences, overlooking long-distance non-linear temporal relations and reversed motions among frames. Such deficiencies undeviatingly result in a nonnegligible performance loss for video action recognition: even a simple action, such as ``sit'' shown in Figure \ref{fig:1}, can not be correctly recognized.

To address the above issues, we introduce a Temporal Transformer network with Self-supervision (TTSN) to efficiently and accurately perform complex temporal motion information modeling. Our TTSN consists of an Efficient Temporal Transformer module (ETT) combined with a Temporal Sequence Self-supervision (TSS) module. Regarding the temporal transformer module, it mainly includes a time position embedding sub-module and a time transformer coding layer sub-module, which is used to model complex long-range non-linear motion information. Furthermore, the ETT module is able to: (1) perform pixel-level modeling on the temporal dimension of a given video; (2) excavate motion information representations held in non-local frames; (3) recognize and transmit motion-sensitive pixels in a given spatial dimension.

Naturally, human beings, after iterated learning, possess the spontaneous adeptness to define complex, even befuddling, actions, such as ``sit'' and ``stand'', which hold high spatial similarity yet reverse to one another in temporal sequence order. Spurred by this rationale, we harness self-supervision learning to model reverse motion learning and devise a Temporal Sequence Self-supervision module. The efficient temporal transformer module learns an action from front to back, while the temporal sequence self-supervision module learns the very one over from back to front. Thus, the network can learn and recognize the variability of motions of complex actions in any given temporal dimension, thanks to which the network is capable of distinguishing these actions correctly and efficiently. In addition, the temporal sequence self-supervision module utilizes unlabeled data to let a neural network robustly learn global reverse motion information representations in a given temporal dimension, which is manageable, ingenious, versatile, and smooth to integrate into a training phase of a network with only a thin volume of supplementarily added computational overheads.

\begin{figure*}[tp!]
    \centering
    \includegraphics[width=1\linewidth]{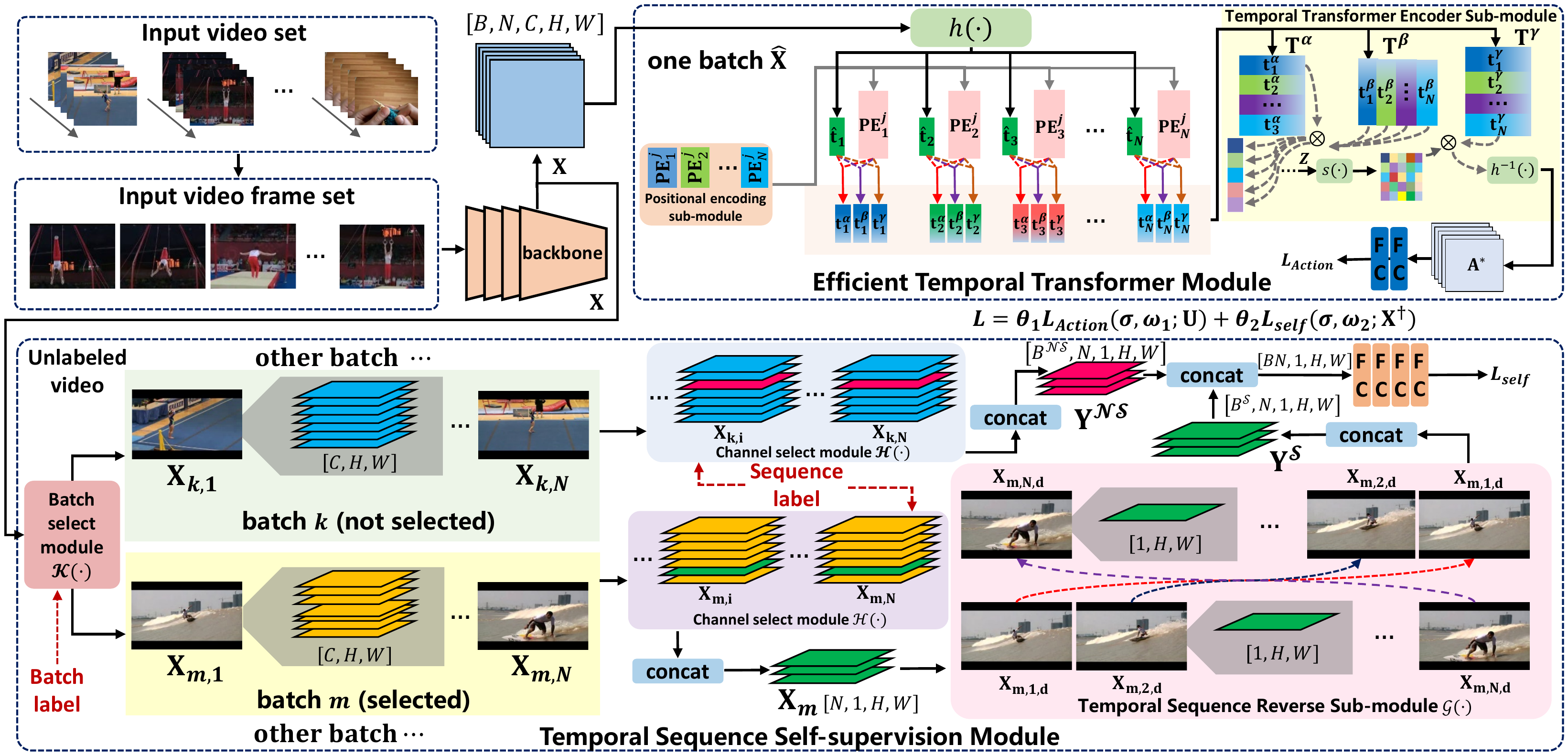}\\
     \caption{The overall architecture of our TTSN. Our TTSN consists of a Backbone $\mathcal{F}$, a Efficient Temporal Transformer Module (ETT), and a Temporal Sequence Self-Supervision Module (TSS). In the training phase, we employ both the Efficient Temporal Transformer module and the Temporal Sequence Self-supervision module; in the testing phase, we only employ the Efficient Temporal Transformer module; in the Efficient Temporal Transformer module, we take one batch of videos only as a manageable instance.}
    \label{fig:2}
\end{figure*}

Foremostly, our contributions can be summarized as follows:

1) To our best knowledge, we unprecedentedly introduce an innovative 2D CNNs-based action recognition network, Temporal Transformer Network with Self-supervision (TTSN). Our TTSN abandons unnecessarily copious amounts of optical flow computation and thanks to which it can outmatch the performance of 3D CNNs while still managing to preserve the complexity of 2D CNNs.

2) We introduce a high-performance temporal transformer module to excavate long-range and complex non-linear motion information representations carried in a temporal dimension and perform pixel-level dependency modeling.

3) We devise a temporal sequence self-supervision module to let the network learn actions from back to front along a temporal dimension, applied by which the network can extract a thoroughness of high-level motion direction perception information to enhance the robustness of the network and recognize complex actions.

4) Our proposed TTS model, compared to the existing state-of-the-art methods, achieves the most high-grade performance on several mainstream datasets, such as HMDB51, UCF101, and Something-Something V1, demonstrated by a large number of extensive experiments conducted.

\section{Related Work}

In this section, we review several of the most related methods for 2D CNNs-based Action Recognition, Self-Attention, and Self-Supervised Learning.

\subsection{2D CNNs-based Action Recognition}

Recently, video action recognition is a prominent research hotspot. Although 3D-CNNs methods perform well, they are computationally costly and laborious to deploy and incrementally develop. With the recent development of deep neural networks, 2D CNNs-based methods are prevalent among video action recognition solutions. The work~\cite{2018TSM} proposes a time shift module to move part of the channel along the temporal dimension to model the motion information among adjacent frames. And the work~\cite{2019STM} proposes a spatio-temporal coding module to encode and model spatio-temporal features. In considering computational performance, a temporal information enhancement module~\cite{2019TEINet} is introduced to decouple the correlation among channels and interactively model the temporal information.~\cite{2020Temporal} proposes a frame-level feature filtering mechanism to diminish information redundancy.~\cite{2020TEA} designs a time excitation and aggregation module to calculate motion-sensitive channels.~\cite{2019Interlaced} proposes a time pyramid network to perform multi-scale modeling of spatio-temporal features.~\cite{2020TAM} proposes an efficient temporal dynamic convolution kernel. In addition, a method,~\cite{2020TDN}, is worth discussing, using the difference information among frames to model.~\cite{2021ACTION} performs adaptive calculations on spatio-temporal characteristic and channel response.

\subsection{Self-Attention}
The author of the self-attention mechanism first proposed this method in natural language processing. Gradually, this method permeates the research of computer vision. The self-attention mechanism~\cite{8101020,8943103,8955791} allows the neural network to selectively focus on a particular part of the input and extract useful neural network perception information. Vaswani’s groundbreaking work~\cite{2017Attention} proposed the self-attention mechanism for the first time and successfully applied it to natural language processing solutions. Since then, researchers of computer vision have also adapted this method.~\cite{2020An} unites the self-attention mechanism with the transformer network structure and applies it to a solution of image classification.~\cite{2018Interaction} proposes an interactive self-attention model to model pixels in a local space.~\cite{2020Dual,2018OCNet} employs cross-sparse self-attention to model the feature map in a spatial perspective.~\cite{2017Non} performs simple pixel-level modeling of non-local feature maps.

\subsection{Self-supervised Learning}
\noindent Self-supervised learning is a novel neural network learning model. It practices an auxiliary pretext task to provide supervision signals for feature learning using unlabeled training data, thanks to which the network can learn the transferable visual feature representation and apply the representation to downstream tasks. This practice also improves the performance and enhances the omnipresence of the network. Traditional self-supervised learning excavates transferable visual information representation from the spatial dimension.~\cite{gidaris2019boosting} practices a rotation pretext task, which randomly rotates the input image. The classifier is employed to control the rotation angle of the image.~\cite{2016Colorful} practices the pretext task of predicting the color of a specific channel.~\cite{2016Context} proposes the pretext task of image restoration from the perspective of spatial context. After developing rapidly, self-supervised learning has also been frequently adopted in other particular fields of computer vision, such as medical image segmentation~\cite{2020Self}, unsupervised visual representation learning~\cite{2015Unsupervised, 2016Unsupervised, 2019Momentum}, knowledge distillation~\cite{2020Knowledge}, re-identification~\cite{2020CycAs}, unsupervised optical flow algorithm~\cite{2020What}, and dense face alignment~\cite{RN1017,2020,RN997}.

\section{Temporal Transformer Network with Self-supervision}

In this section, we first present an overview of our method, including the crucial components of our proposed TTSN and a concise description of each essential module. We then describe each module specifically, including the precise principle we utilize, implementation details, formalized definitions, etc. Eventually, we present a formal definition of the loss function and elucidate its rationale.

\subsection{Overview}

In this paper, we introduce a novel method, Temporal Transformer Network with Self-supervision (TTSN), a 2D CNNs-based video action recognition network, for video action recognition. Our TTSN principally consists of an Efficient Temporal Transformer (ETT) module, a Temporal Sequence Self-supervision (TSS) module, and a ResNet-based backbone. The Efficient Temporal Transformer module is chiefly engaged in modeling, both non-local and non-linear, of a given temporal dimension and thoroughly excavating the motion information carried in the temporal dimension of a given video. The Temporal Sequence Self-supervision module is for reverse-ordered processing of unlabeled, randomly sampled videos from a dataset. When executing the reverse-ordered processing, the TSS module randomly selects a channel to process. Ultimately, the TSS module necessitates the backbone to percept and derive rich and high-level motion direction information; it also enhances the robustness of the network, making it more sophisticated in distinguishing complex actions.

As shown in Figure 2, the input video set contains multiple variable-length videos. The sparse temporal sampling strategy proposed in~\cite{2016Temporal,2020TDN} is employed to obtain an extensive number of video framesets with $N$ frames for each video. In the phase of feature extraction, we assign $B$ videos as the input, where $B$ is the size of a batch. Input videos initially pass through the backbone and collect a feature map $\bm{\mathrm{X}}\in \mathbb{R}^{B\times N\times C\times H \times W}$.  Distinctively, in the phase of training, $\bm{\mathrm{X}}$ passes through the Efficient Temporal Transformer module and the Temporal Sequence Self-supervision module; while in the phase of testing, it passes exclusively through the Efficient Temporal Transformer module and completes an efficient inference. Eventually, we utilize the corresponding loss function to constrain the output of ETT and TSS to reach a total loss, after which a robust non-linear motion modeling in both directions, forward and backward, can be realized.

\subsection{Efficient Temporal Transformer Module}
A temporal dimension of a given video contains rich motion information. Therefore, the ability of a network to model motion information would be essential to the model and determine the performance of the final classification. Inspired by~\cite{2017Attention,2020An}, we propose a simple temporal transformer structure, namely an Efficient Temporal Transformer module, to model the complex motion information of a temporal dimension with a non-linear and non-local method.

According to Figure \ref{fig:2}, the Efficient Temporal Transformer module accepts feature map $\bm{\mathrm{X}}$ as input and principally comprises a positional encoding sub-module and a temporal transformer encoder sub-module. To demonstrate explicitly, here we take only one batch (batch $k$) of $\bm{\mathrm{X}}$ as an instance and reasonably redefine $\bm{\mathrm{X}}_k$ as $\hat{\bm{\mathrm{X}}}=\{\hat{\bm{\mathrm{x}}}_i\in \mathbb{R}^{C\times H\times W}|i=\{1,2,...,N\}\}$ to elucidate the principle of the Efficient Temporal Transformer module. In enhancing the modeling ability of the ETT module and avoiding long-winded computation, a linear transformation function $h(\cdot)$ is employed to accomplish frame-level feature embedding and is tied for the transformation of the feature map $\hat{\bm{\mathrm{x}}}_i$, altering the dimension of each video frame from $C\times H\times W$ to $1\times l$ ($l\ll CHW$), and obtaining the 1D frameset $\hat{\bm{\mathrm{T}}}=\{\hat{\bm{\mathrm{t}}}_i\in\mathbb{R}^{1\times l}|i=\{1,2,...,N\}\}$. Subsequently, we add a learnable and randomly initialized positional encoding ${\bm{\mathrm{PE}}}^j_i\in\mathbb{R}^{1\times l}$ ($j\in \{\alpha,\beta,\gamma\}$) to $\hat{\bm{\mathrm{t}}}_i$, representing the raw 1D frame tensor forsaking the positional encoding obtained after each frame $\hat{\bm{\mathrm{x}}}_i$ passes through the frame embedding module $h(\cdot)$. The above process can be formalized as follows: 

\begin{equation}\label{ett-1}
	{\bm{\mathrm{t}}}^j_i = \hat{{\bm{\mathrm{t}}}}_i+ \bm{\mathrm{PE}}^j_i,
\end{equation}
\begin{equation}\label{ett-2}
	\hat{\bm{\mathrm{t}}}_i=h(\hat{\bm{\mathrm{x}}}_i),
\end{equation}

\noindent where ${\rm {\rm \textbf{t}}}^j_i$ is the 1D frame tensor with positional encoding. 

To facilitate the Temporal Transformer Encoder sub-module to learn temporal non-linear motion features more versatilely and convertibly, we add three sets of independent positional encodings to each 1D frame tensor and collect the calculation result of each frame ${\bm{\mathrm{t}}}^j_i$, as shown in Equation (\ref{ett-1}). Subsequently, we sequentially concatenate all 1D frame tensors ${\rm {\rm \textbf{t}}}^j_i$ to respectively construct three independent temporal encoding feature maps $\textbf{T}^j$, where $j\in \{\alpha,\beta,\gamma\}$.

As for the temporal transformer encoder, we use ${\rm \textbf{T}}^{\alpha},{\rm \textbf{T}}^{\beta},{\rm \textbf{T}}^{\gamma}$ to process non-linear and non-local modeling on a temporal dimension of a given video. We devise a simple and efficient temporal transformer encoder sub-module, containing no MLP layer and other unnecessary structures such as the decoder structure. It not only lessens the computational cost and reasoning time consumption but also promotes the modeling ability of the model in the temporal dimension. The rationale of the transformer encoder can be formalized as follows:

\begin{equation}\label{1-1}
	\bm{\mathrm{A^*}}=h^{-1}(s(\bm{\mathrm{Z}}){\bm{\mathrm{T}}^{\gamma}}),
\end{equation}
\begin{equation}\label{1-2}
	\bm{\mathrm{Z}}=\lambda{\bm{\mathrm{T}}}^{\alpha}({\bm{\mathrm{T}}}^{\beta})^{T},
\end{equation}

\noindent where $s(\cdot)$ is the normalization function. ${\rm \textbf{A}}^{*}$ represents the auxiliary attention map with dimension $BN\times C\times H\times W$, containing motion sensitive pixel information in the temporal dimension. These motion-sensitive information can guide the network to complete the classification of action categories. $h^{-1}(\cdot)$ represents the operation of inverse transformation. $\bm{\mathrm{Z}}$ represents the relation matrix, portraying the long-range non-linear dependencies of frames along the given temporal dimension. $\lambda$ is a learnable parameter and is used to enhance the learning ability and improve the expressiveness of the network.

We consider that $\textbf{A}^{*}$ is capable of guiding the network to complete the classification of action categories since it concentrates on the region of motion-sensitive pixels on the spatial scale of video frames, supported by the notion that regions where motion-sensitive pixels are exactly where the network's attention should be. The attention is calculated by the Temporal transformer encoder sub-layer in the Efficient Temporal Transformer module through non-local and non-linear modeling on the given temporal dimension. In the experiment section, we conduct a visualization operation on ${\rm \textbf{A}}^{*}$.

Here we succinctly elucidate the implementation of each sub-module of the Efficient Temporal Transformer module. Firstly, $h(\cdot)$ represents the frame-level feature embedding sub-module and can be realized by a convolution and a linear projection operation. The positional encoding sub-module can generate a learnable and randomly initialized positional encoding, which is superimposed directly on 1D frame tensor $\hat{{\rm \textbf{t}}}_i$. Next, $h^{-1}(\cdot)$ represents the inverse transformation sub-module, which can be realized by a convolution operation. And $s(\cdot)$ is a normalization function implemented with Softmax.

\subsection{Temporal Sequence Self-Supervision Module}
Human beings learn a new action by repeatedly observing; regularly, this process acquires a large set of instances, especially when the one is complex and befuddling. These actions, yet reversed along the temporal sequence, are similar in the spatial dimension. When a temporal sequence is modified or changed, actions on the sequence may be mistreated as different ones. Some of these actions are too complicated for 2D CNNs-based methods \cite{2020TDN} to process; therefore, when the model can distinguish this temporal difference, the classification performance, as well as the robustness of the model, could be enhanced. Inspired by the learning behavior in human beings, we examine self-supervised learning to model reversed motion learning with a Temporal Sequence Self-supervised (TSS) module with unlabeled video data, which can incorporate with the ETT module to achieve bidirectional motion information modeling. Concretely, we consider processing the actions in reversed order along the temporal dimension and contrive to make the neural network learn to distinguish whether the current action is in the reversed order along the temporal dimension. We endeavor to enable the model to learn the same action from back to front along the temporal dimension. In this way, the model can distinguish the complicated actions stated above; also, the robustness of the network is strengthened.

To achieve the best performance when assigning auxiliary tasks to the Temporal Sequence Self-supervision module, we devise a myriad of algorithms, namely ``All batch all channel-Reverse ($\mathcal{AA}$)'', ``Random batch all channel-Reverse ($\mathcal{RA}$)'', ``All batch random channel-Reverse ($\mathcal{AR}$)'', and ``Random batch random channel-Reverse ($\mathcal{RR}$)''. 

$\mathcal{RR}$ algorithm possesses the highest randomness and leading performance among the others; so in TTSN, we use the algorithm of $\mathcal{RR}$ by default. As shown in Figure \ref{fig:2}, after inputting the feature map into the Temporal Sequence Self-Supervision module, the $\mathcal{RR}$ algorithm first generates batch-level pseudo-labels and sequence-level pseudo-labels for the input feature maps. For batch-level pseudo-labels, there are two categories: Selected Batch (termed as $\mathcal{S}$) and Not Selected Batches (termed as $\mathcal{NS}$). For sequence-level pseudo-labels, there are also two categories: Temporal Norm Sequence (termed as $NOR$) and Temporal Reverse Sequence (termed as $REV$). 

For batches (e.g., batch $k$) whose label is $\mathcal{NS}$, we randomly select one channel (e.g., channel $d_1$) from each frame in the batch and maintain the original temporal sequence unchanged, then the feature maps ${\bm{\mathrm{X}}}_{k,d_1}\in \mathbb{R}^{N\times 1\times H\times W}$ can be obtained. Finally, we concatenate them to construct the result feature maps $\bm{\mathrm{Y}}^{\mathcal{NS}}\in {\mathbb{R}^{B^{\mathcal{NS}}\times N\times 1\times H\times W}}$, where $B^{\mathcal{NS}}$ is the number of unselected batches; in addition, the sequence-level pseudo-label of the video frames in the batch is $NOR$ to form the label set $\bm{\mathrm{G}}^{\mathcal{NS}}\in{\mathbb{R}^{B^{\mathcal{NS}}}}$. For batches (e.g., batch $m$) with a pseudo-label of $\mathcal{S}$, we also randomly select one channel (e.g., channel $d_2$) from the $N$ frames in the batch to obtain the feature maps ${\bm{\mathrm{X}}}_{m,d_2}$. Then we reverse the order of the channels in the temporal dimension to obtain the feature maps $\bm{\mathrm{Y}}^{\mathcal{S}}\in {\mathbb{R}^{B^{\mathcal{S}}\times 1\times H\times W}}$, where $B^{\mathcal{S}}$ is the number of the selected batches. Eventually, the sequence-level pseudo-label of the video frame is $REV$, forming the label set $\bm{\mathrm{G}}^{\mathcal{S}}\in{\mathbb{R}^{B^{\mathcal{S}}}}$.

\begin{table}[tp!]
    \centering
    \caption{Relation between the self-supervision algorithm and the different values of hyper-parameters $\rho,\eta$.}
    \fontsize{9}{11}\selectfont    
    \begin{tabular}{c|c|c|c}
    \hline
    Algorithm & $\Gamma$ & $\rho$ & $\eta$\\  
    \hline
      $\mathcal{AA}$ &$\mathcal{G}$ & 1 & 1  \\
      $\mathcal{RA}$ &$\mathcal{G}\circ\mathcal{K}$ & 1 & 0   \\
      $\mathcal{AR}$ &$\mathcal{G}\circ\mathcal{H}$ & 0 & 1 \\
      $\mathcal{RR}$ &$\mathcal{G}\circ\mathcal{H}\circ\mathcal{K}$ & 0 & 0\\
     \hline
    \end{tabular}\vspace{0.2cm}
    \label{tab:tss-diff-v}
\end{table}

$\mathcal{AA}$ selects all batches, reverses all frames, and uses all channels in the reversing operation; $\mathcal{RA}$ randomly selects one batch to reverse and uses all channels; $\mathcal{AR}$ selects all batches to reverse, and randomly selects one channel from each batch to reverse. The rationale of the Temporal Sequence Self-supervision module can be formalized by the follows:

\begin{equation}\label{TSS-1}
    \Gamma\left(\varepsilon,\rho,\eta\right)=\left(1-\varepsilon\right)\mathcal{G}\circ\left(1-\rho\right)\mathcal{H}\circ\left(1-\eta\right)\mathcal{K},
\end{equation}

\noindent where “$\circ$” represents the compound operation of the function; $\mathcal{G}(\cdot)$ represents the temporal sequence reversion; $\mathcal{H}(\cdot)$ represents the random channel selection; $\mathcal{K}(\cdot)$ represents the random batch selection; $\varepsilon,\rho,\eta$ represent the hyper-parameters to control the final self-supervision algorithm $\Gamma$, where $\varepsilon,\rho,\eta
\in\{0,1\}$ and $\Gamma\in\{\mathcal{AA},\mathcal{AR},\mathcal{RA},\mathcal{RR}\}$. Notice that the operations of $\mathcal{G},\mathcal{H}$ and $\mathcal{K}$ do not comply to the commutative law.

\begin{algorithm}[tp!]
\caption{$\mathcal{RR}$ algorithm of the Temporal Sequence Self-Supervision Module}
\label{alg:algorithm}
\textbf{Input}: Training video set $\bm{\mathrm{X}}\in \mathbb{R}^{B\times N\times C\times H\times W}$ and corresponding hyper-parameters $\varepsilon=0,\rho=0,\eta=0$. \\
\textbf{Output}: The output feature maps $\bm{\mathrm{Y}}$ and corresponding labels $\bm{\mathrm{G}}$.\\
% \text{$\qquad \quad \,\,\,\,\,$ \emph{BN} sequence labels}
\begin{algorithmic}[1] %[1] enables line numbers
\STATE{Init $\bm{\mathrm{Y}}$,$\bm{\mathrm{G}}$=\{\}}
\FOR{$b$ in B}
\STATE{Init $\bm{\mathrm{Y}^{\mathcal{NS}}}$,$\bm{\mathrm{Y}^{\mathcal{NS}}}$,$\bm{\mathrm{G}}^{\mathcal{NS}}$,$\bm{\mathrm{G}}^{\mathcal{S}}$=\{\}}
\STATE{$s_b=\mathcal{K}\left(\textbf{X}_b\right)$}
\IF{$s_b=\mathcal{NS}$}
\STATE{$\bm{\mathrm{G}}_{b,{\forall}{N}}=NOR$}
\STATE{$d_1=\mathcal{H}({\textbf{X}}_b)$}
\STATE{$\bm{\mathrm{Y}}_{b}=\textbf{X}_{b,d_1}$}
\STATE{$\Leftrightarrow[\textbf{x}_{b,1,d_1},\textbf{x}_{b,2,d_1},...,\textbf{x}_{b,N,d_1}]$}
\STATE{$\bm{\mathrm{Y}}^{\mathcal{NS}}$=[$\bm{\mathrm{Y}}^{\mathcal{NS}}$,$\bm{\mathrm{Y}}_{b}$],$\bm{\mathrm{G}}^{\mathcal{NS}}$=[$\bm{\mathrm{G}}^{\mathcal{NS}}$,$\bm{\mathrm{G}}_{b,{\forall}{N}}$]}
\ELSIF{$s_b=\mathcal{S}$}
\STATE{$\bm{\mathrm{G}}_{b,{\forall}{N}}=REV$}
\STATE{$d_2=\mathcal{H}({\textbf{X}}_b)$}
\STATE{$\bm{\mathrm{Y}}_{b}=\mathcal{G}(\textbf{X}_{b,d_2})$}
\STATE{$\Leftrightarrow[\textbf{x}_{b,N,d_2},\textbf{x}_{b,N-1,d_2},...,\textbf{x}_{b,1,d_2}]$}
\STATE{$\bm{\mathrm{Y}}^{\mathcal{S}}$=[$\bm{\mathrm{Y}}^{\mathcal{S}}$,$\bm{\mathrm{Y}}_{b}$],$\bm{\mathrm{G}}^{\mathcal{S}}$=[$\bm{\mathrm{G}}^{\mathcal{S}}$,$\bm{\mathrm{G}}_{b,{\forall}{N}}$]}
\ENDIF
\STATE{$\bm{\mathrm{Y}}$=[$\bm{\mathrm{Y}}$,$\bm{\mathrm{Y}}^{\mathcal{NS}}$,$\bm{\mathrm{Y}}^{\mathcal{S}}$], $\bm{\mathrm{G}}$=[$\bm{\mathrm{G}}$,$\bm{\mathrm{G}}^{\mathcal{NS}}$, $\bm{\mathrm{G}}^{\mathcal{S}}$]}
% $\bm{\mathrm{G}}^{\mathcal{S}}$=[$\bm{\mathrm{G}}^{\mathcal{S}}$,$\bm{\mathrm{G}}_{b,{\forall}{N}}$]}
\ENDFOR
% \STATE{return $\textbf{output}\in\mathbb{R}^{BN\times 1\times H\times W}$}
\end{algorithmic}
\end{algorithm}

Based on different values of $\varepsilon,\rho,\eta$ in formula (\ref{TSS-1}), $\Gamma$ can be transformed into different self-supervision algorithms. The coefficients of the function (e.g., $\mathcal{G}(\cdot),\mathcal{H}(\cdot)$ and $\mathcal{K}(\cdot)$) remains 0 if the function does not participate in the compound operation of function, otherwise 1.
Notice the reversion can only execute for the TTSN model, so we assign $\varepsilon$ to 0. The relation between the self-supervision algorithm and different values of hyper-parameters $\rho,\eta$ is manifested in Table \ref{tab:tss-diff-v}.

Since the algorithm of $\mathcal{RR}$ demands the most moderate GPU configuration notwithstanding still performs the best (this will be demonstrated more evidently in the experiment section), we select $\mathcal{RR}$ as the auxiliary task for the Temporal Sequence Self-supervision module in this paper. We take the $\mathcal{RR}$ algorithm as an instance to illustrate the fundamental flow, which is detailed in Algorithm \ref{alg:algorithm}.

\subsection{Loss Function}
In this section, we elucidate the loss function of our TTSN. It consists of two components: an Efficient Temporal Transformer module loss $L_{Action}$ and a Temporal Sequence Self-supervision loss $L_{self}$. The $L_{Action}$ is for constraining the efficient temporal transformer module to model complex non-linear motion information in a given temporal dimension and ultimately perform accurate classification. Likewise, the $L_{self}$ is for constraining the backbone and deriving reversed motion features.

\noindent\subsubsection{Loss Function of ETT} The Efficient Temporal Transformer module we proposed is very adaptable, which can be added to veritably any locality of the network. Supported by our experience, with an added locality getting imminent to the very front of the network, the demand for extra computational overheads induced about would expand exponentially, causing a considerable loss in the performance; such an impairment is uninvited. Therefore, to ensure efficiency and accuracy in computation, we append the ETT module to the end of our network to excavate the underutilized motion information in the given temporal dimension of a high-level abstract feature map. In the process of training, the loss function of the Efficient Temporal Transformer is defined as follows:

\begin{equation}\label{loss:1}
L_{Action}\left(\sigma,{\omega}_1;{\rm \textbf{U}}\right)=-\sum\limits_{p=1}^{c}{t_{p}log\left(\mathcal{R}_{{\omega}_1}^{p}\left({\bm{\mathrm{A}}}^{*}+\bm{\mathrm{X}}\right)\right)},
\end{equation}

\noindent where $c$ represents the total number of categories; $t_p$ represents the category label; $\mathcal{R}_{\rho}^{p}\left(\cdot \right)$ represents the standardized prediction score of action category $p$; $\sigma$ represents the parameters of the backbone; ${\omega}_1$ represents the parameters of the action classifier. ${\rm \textbf{U}}$ is the whole training set of the corresponding dataset.

\noindent\subsubsection{Loss Function of TSS} During the training process, we implement an auxiliary self-supervised loss function to constrain the training process of TTSN in the temporal dimensions. In this work, the training set ${\rm \textbf{U}^{\dagger}}$ is from the parental training set ${\rm \textbf{U}}$; modifications are that we remove its original labels and introduce new pseudo-labels to grant supervision signals. The loss function of the Temporal Sequence Self-Supervision module is defined as follows:

\begin{equation}\label{loss:2}
\begin{aligned}
L_{self}(\sigma,\omega_2;{\rm \textbf{U}^\dagger})= -\sum\limits_{{t_q\in\{0,1\}}}^{}\left(t_qlog\left(\mathcal{H}^q_{{\omega}_2}\left([\bm{\mathrm{Y}}^{\mathcal{NS}},\bm{\mathrm{Y}}^{\mathcal{S}}]\right)\right)\right),
\end{aligned}
\end{equation}

\noindent where $t_q\in\{0,1\}$ represents the value of pseudo-labels of selected and unselected batches; ${\bm{\mathrm{U}}}^{\dagger}\subseteq {\bm{\mathrm{U}}}$ represents an unlabeled dataset randomly sampled from the training set ${\rm\textbf{U}}$, whose original labels have been deprecated; $\mathcal{H}^q_{\omega_2}\left(\cdot \right)$ represents, with respect to the prediction of the temporal order of frames, the normalized prediction scores of the selected and unselected batches; $\omega_2$ represents the parameters of the self-supervised classifier.

\noindent\subsubsection{Total Loss of TTSN} Combining Formulation (\ref{loss:1}) and (\ref{loss:2}), we define the total loss function as follows:
\begin{equation}\label{loss:3}
L=\theta_1 L_{Action}\left(\sigma,\omega_1;{\rm \textbf{U}}\right)+\theta_2 L_{self}\left(\sigma,\omega_2;{\rm\textbf{U}}^{\dagger}\right),
\end{equation}
\noindent where $\theta_1$ and $\theta_2$ are hyper-parameters used to balance the weights of the ETT module and the respective TSS module.

According to the experiments we conducted, being the value of $\theta_2$ grows, concurrently, the confinement $\mathcal{E}$ of the Temporal Sequence Self-supervision module strengthens, which satisfies $\mathcal{E}\propto \theta_2$. We assign $\theta_1=1.0$ and restrict $\theta_2$ to qualify the relation $\theta_1 / \theta_2 \in \left[10,100\right]$ empirically; the value of $\theta_2$ depends on the scale of the dataset.

\section{Experiment}
In this section, we first establish the context and settings of our experiments; subsequently, we present the results of our proposed TTSN model being experimented on three frequently-used datasets: HMDB51, UCF101, and Something-Something V1 and compare them with several state-of-the-art methods. This section culminates in the exhibition of our detailed ablation experiments and visualized results for each.

\subsection{Datasets}
We verify our proposed TTSN on three frequently-used datasets: HMDB51, UCF101, and Something-Something V1. Among them, HMDB51~\cite{2011HMDB} contains 51 categories and 6849 videos. Each category comprises at least 51 videos, most of which are from YouTube and Google, with a resolution of 320$\times$240. UCF101~\cite{2012UCF101} is a dataset particularly for action recognition, comprising numerous real action videos which are from YouTube, providing 13320 videos from 101 categories. Something-something V1~\cite{8237884} is a relatively massive dataset of 174 categories of actions and 108499 videos, containing various daily actions between humans and mundane objects. It would be unlikely to determine what the individual in a video is doing depending solely on one frame extracted. The above three datasets comprehensively evaluate our proposed TTSN model from different regards and viewpoints.

\subsection{Implementation Details}
We employ ResNet50 as the backbone to realize our proposed TTSN. We apply $8f$ and $16f$ settings which signify we extract $N$ frames from each video. TTSN uses these $N$ frames in each video for modeling; we assign $N=8$ and $N=16$. In the pre-training stage of TTSN, conventionally, we adopt two different pre-train strategies that ImageNet solitarily or ImageNet + Kinetics combined are employed. As for the efficient temporal transformer module, we employ only one temporal transformer encoder sub-module to lessen reasoning time and improve the performance. In the temporal sequence self-supervision module, we employ “Random batch random channel-Reverse” as this algorithm performs the best and holds the strongest randomness. In succeeding ablation experiments, we also trial on other algorithms we propose, including $(\mathcal{AA}, \mathcal{RA}, \mathcal{AR})$, for the temporal sequence self-supervision module.

Since the three datasets contain distinct video quantities, we apply different training strategies and hyper-parameter settings to each dataset. For HMDB51 and UCF101 datasets, we assign $lr = 1.5\times10^{-4}$ and train our TTSN for 50 epochs with $B=4$. For the Something-Something V1 dataset, we assign $lr = 1.3\times10^{-3}$ and train our TTSN for 65 epochs with $B=8$. If the verification set performance saturates, the learning rate would be divided by 10. In this paper, we set $lr\_steps=\left[25,35\right]$ for HMDB51 and UCF101, and $lr\_steps=\left[30,45,55\right]$ for Something-something V1. Training and testing employ a center crop of $224\times224$ from a single clip. From HMDB51, UCF101 to Something-Something V1, the scale of the dataset is increasing. The temporal sequence self-supervision module demands a stronger constraint ability to deal with a large-scale dataset so that the network can learn motion information representations robustly. Therefore, for HMDB51, we empirically 
specify $ (\theta_1=1.0,\theta_2=0.01)$ in loss function Formulation (\ref{loss:3}). For UCF101 and Something-Something V1, we specify $(\theta_1=1.0,\theta_2=0.1)$. 

\begin{table}[tp!]
    \centering
    \caption{Comparison between our TTSN and several of the state-of-the-art methods on HMDB51. ``*'' denotes the result is collected from running the code open-sourced by the method's author.}
    \fontsize{8}{10}\selectfont   
     \resizebox{\linewidth}{!}{
    \begin{tabular}{c|c|c|c}
    \hline
    Method & Pretrain & Backbone & Top1.(\%)\\
\hline
    C3D \cite{2014Learning} & Sports-1M & ResNet18 & 54.9 \\
    TSN \cite{2016Temporal} & ImageNet & Inception V2 & 53.7 \\
    TDN* \cite{2020TDN} & ImageNet & ResNet50 & 56.4 \\
    \hline
    \textbf{TTSN} & ImageNet & ResNet50 & \textbf{60.2}\\
    \hline
    ARTNet \cite{2017Appearance} & Kinetics & ResNet18 & 70.9 \\
    TSM \cite{2018TSM} & Kinetics & ResNet50 & 73.2 \\    
    R(2+1)D \cite{0A} & Kinetics & ResNet34 & 74.5 \\
    STM \cite{2019STM} & ImageNet+Kinetics & ResNet50 & 72.2 \\
    TEA \cite{2020TEA} & ImageNet+Kinetics & ResNet50 & 73.3 \\
    I3D \cite{2017Quo} & ImageNet+Kinetics & Inception V2 & 74.8\\
    S3D \cite{2017Rethinking} & ImageNet+Kinetics & Inception V2 & 75.9 \\
    TDN* \cite{2020TDN} & ImageNet+Kinetics & ResNet50 & 79.1\\
    \hline
    \textbf{TTSN} & ImageNet+Kinetics & ResNet50 & \textbf{80.2}\\
    \hline
    \end{tabular}}\vspace{0.2cm}
    \label{tab:hmdb51-sota}
\end{table}

\subsection{Comparison with the state-of-the-art}
In this section, we respectively report the results of our TTSN model on HMDB51, UCF101, and Something-Something V1 datasets. When experimenting, We instantiate our proposed TTSN with the backbone of ResNet50 and compare it to other state-of-the-art methods with a similar backbone. Firstly, we thoroughly verify our proposed TTSN on the HMDB51 dataset, as is shown in Table \ref{tab:hmdb51-sota}. From the table, we can see that when our TTSN loads only the pre-trained model on ImageNet, $Top1.$ reaches 60.2\%. To the best of our knowledge, compared with the latest 2D CNN-based action recognition method TDN, the result improves by $3.8\%$. More importantly, it should be emphasized that our ImageNet pre-trained model outperforms some Sport-1M pre-trained 3D CNN models (e.g., C3D~\cite{2014Learning}) in recognition accuracy. When our TTSN loads the pre-trained model on Image + Kinetics combined, $Top1.$ reaches 80.2\%, the best performance of the model based on 2D CNNs on HMDB-51, even if compared with the latest TDN model.

\begin{table}[tp!]
    \centering
    \caption{Comparison between our TTSN and several of the state-of-the-art methods on UCF101. ``*'' denotes the result is collected from running the code open-sourced by the method's author.}
    \fontsize{8}{10}\selectfont  
     \resizebox{\linewidth}{!}{
    \begin{tabular}{c|c|c|c}
    \hline
    Method & Pretrain & Backbone & Top1.(\%)\\
    \hline
    C3D \cite{2014Learning} & ImageNet & ResNet18 V2 & 85.8 \\
    TSN \cite{2016Temporal} & ImageNet & Inception V2 & 86.4 \\
    TDN* \cite{2020TDN} & ImageNet & ResNet50 & 85.8\\ 
    \hline
    \textbf{TTSN} & ImageNet & ResNet50 & \textbf{86.4}\\
    \hline 
    STC \cite{2018Spatio} & Kinetics & ResNet101 & 93.7\\ 
    TSM \cite{2018TSM} & Kinetics & ResNet50 & 96.0 \\
    ARTNet \cite{2017Appearance} & Kinetics & ResNet18 & 94.3\\
    R(2+1)D \cite{0A} & Kinetics & ResNet34 & 96.8\\ 
    StNet \cite{2018StNet} & ImageNet + Kinetics & ResNet50 & 93.5 \\
    I3D \cite{2017Quo} & ImageNet + Kinetics & Inception V2 & 95.6\\
    S3D \cite{2017Rethinking} & ImageNet + Kinetics & Inception V2 & 96.8 \\
    STM \cite{2019STM} & ImageNet + Kinetics & ResNet50 & 96.2 \\ 
    TDN* \cite{2020TDN} & ImageNet + Kinetics & ResNet50 & 96.4 \\
    \hline
    \textbf{TTSN} & ImageNet + Kinetics & ResNet50 & \textbf{96.8}\\
    \hline
    \end{tabular}}
    % \vspace{0.2cm}
    \label{tab:ucf101-sota}
\end{table}

In order to evaluate our proposed TTSN model more, We employ ResNet50 as the backbone and make a comparison with other existing state-of-the-art models on UCF101 dataset. When TTSN only loads the pre-trained model on the ImageNet dataset, $Top1.$ reaches 86.4\%, as is shown in Table \ref{tab:ucf101-sota}. Furthermore, when TTSN loads the pre-trained model on ImageNet + Kinetics combined, $Top1.$ reaches 96.8\%. From the Table we can see that the TTSN model achieves the best recognition results, outmatching some classical 3D CNN models (e.g., C3D~\cite{2014Learning}, P3D~\cite{qiu2017learning}, I3D~\cite{2017Quo}, S3D~\cite{2017Rethinking}) and even several of those with the more powerful ResNet101 as their backbone (e.g., STC ~\cite{2018Spatio}).

To verify our TTSN thoroughly, we also conduct experiments on a larger action recognition dataset Something-Something V1. As shown in Table \ref{tab:something-sota}, we also present the results of our TTSN model under the settings of $8f$ and $16f$. As we can see from the table, under the $8f$ setting, $Top1.$ and $Top5.$ reach 52.4\% and 80.6\% respectively; while under the $16f$ setting, $Top1.$ and $Top5.$ reach 53.3\% and 81.5\%. The Something-Something V1 dataset holds a strong temporal correlation, making it unlikely to accurately determine categories of actions solely based on a single frame. The comparison of our TTSN with the state-of-the-art methods on the Something-Something V1 dataset proves that our TTSN has a powerful temporal modeling capability and performs the best in action recognition.

\begin{table}[tp!]
    \centering
    \caption{Comparison between our TTSN and several of the state-of-the-art methods on Something-Something v1. ``*'' denotes the result is collected from running the code open-sourced by the method's author, while ``-'' denotes unavailable data.}
    \fontsize{12}{14}\selectfont  
    \resizebox{\linewidth}{!}{
    \begin{tabular}{c|c|c|c|c}
    \hline
    Method & Backbone & Frames & Top1.(\%) & Top5.(\%)\\
    \hline
    TSN-RGB \cite{2016Temporal} & BNInception & 8 & 19.5 & - \\
    TRN-Multiscale \cite{2017Temporal} & BNInception & 8 & 34.4 & -  \\
    ECO$_{EN}$Lite \cite{2018ECO} & BN+R18 & 92 & 46.4 & - \\
    S3D-G \cite{2017Rethinking} & Inception & 64 & 48.2 & 78.7  \\
    I3D \cite{2017Quo} & ResNet50 & 32×2 & 41.6 & 72.2\\
    NL I3D+GCN \cite{wang2018videos} & ResNet50+GCN & 32×2 & 46.1 & 76.8 \\
    TAM \cite{2019More} & bLResNet50 & 16×2 & 48.4 & 78.8 \\
    GST \cite{2019Grouped} & ResNet50 & 16 & 48.6 & 77.9 \\
    CorrNet \cite{2019Video} & ResNet50 & 32×10 & 49.3 & - \\
    TSM \cite{2018TSM} & ResNet50 & 8+16 & 49.7 & 78.5 \\
    SmallBigNet \cite{2020SmallBigNet} & ResNet50 & 8+16 & 50.4 & 80.5 \\
    TANet \cite{2020TAM} & ResNet50 & 8+16 & 50.6 & 79.3 \\
    STM \cite{2019STM} & ResNet50 & 16×30 & 50.7 & 80.4 \\
    TEA \cite{2020TEA} & ResNet50 & 16 & 51.9 & 80.3 \\
    TEINet \cite{2019TEINet} & ResNet50 & 8+16 & 52.5 & - \\
    %TDN* \cite{wang2021tdn} & ResNet50 & 8 & 52.3 & 80.6 \\ 
    TDN* \cite{2020TDN} & ResNet50 & 8 & 49.7 & 78.7 \\ 
    TDN* \cite{2020TDN} & ResNet50 & 16 & 52.0 & 80.6 \\
    \hline
    \textbf{TTSN} & ResNet50 & 8 & 52.4 & 80.6\\
    \textbf{TTSN} & ResNet50 & 16 & \textbf{53.5} & \textbf{81.8}\\
    \hline
    \end{tabular}}
    % \vspace{0.2cm}
    \label{tab:something-sota}
\end{table}

\subsection{Ablation Studies}
In this section, we present various ablation studies on the HMDB51 dataset and elucidate the properties of each component of our TTSN, using input settings of $B=4$ and $16f$, and the pre-trained model on the combined ImageNet + Kinetics dataset as initialization.

\begin{figure*}[tp!]
    \centering
    \includegraphics[width=0.9\linewidth]{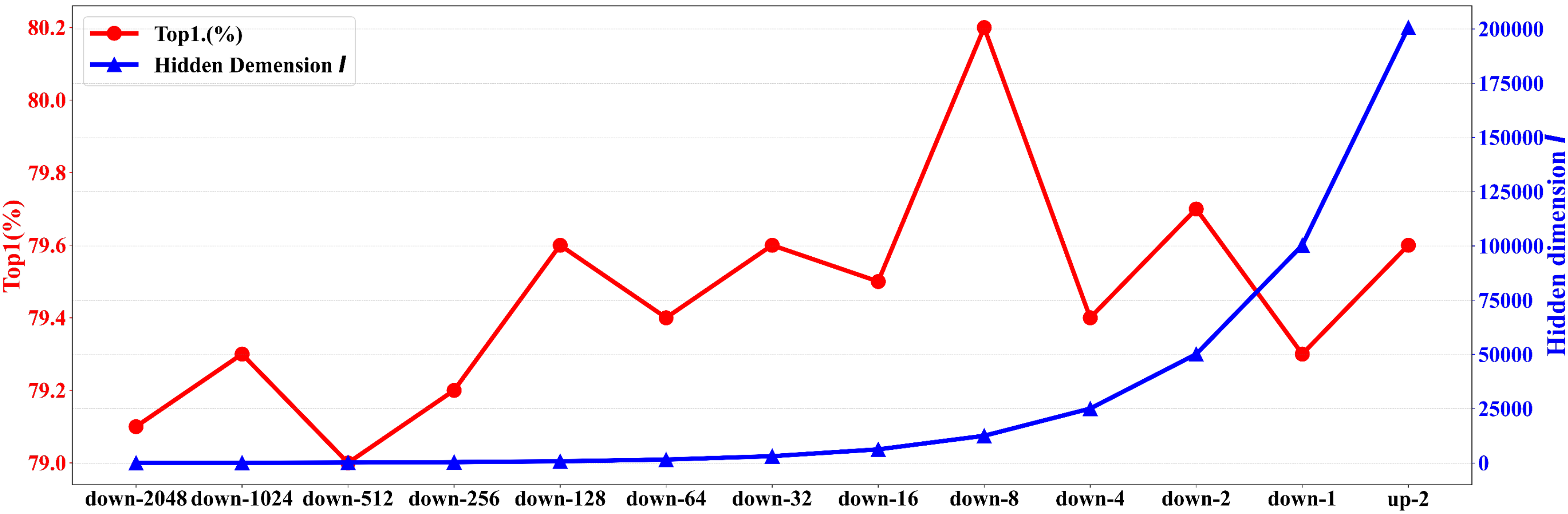}\\
     \caption{Result of the ablation experiment on the hidden dimension $l$ of the temporal transformer encoder sub-module in the Efficient Temporal Transformer Module. The vertical axis on the left side is \textbf{Top1.}(\%), and the vertical axis on the right side is \textbf{Hidden dimension} $l$. The horizontal axis is the channel compression multiplier of $h(\cdot)$ to \textbf{X}. As the channel compression multiplier increases, the hidden dimension $l$ decreases. \textbf{down-}$u$ and \textbf{up-}$u$ denotes compressing and expanding the number of channels of \textbf{X} by $u$ times respectively, and $l\propto\frac{1}{u}C$, $\textbf{X}\in\mathbb{R}^{B\times N\times C\times H\times W}$.}
    \label{fig:3}
\end{figure*}

\noindent\textbf{Complexity Analysis}\,\,We first examine and analyze the time complexity and the number of parameters of the single module proposed in our TTSN. As for the time complexity, we examine the GFLOPs of the Efficient Temporal Transformer module and the Temporal Sequence Self-supervision module. We have also conducted a parameter analysis of these two modules. According to Table \ref{tab:gflops_p}, the Efficient Temporal Transformer module and the Temporal Sequence Self-supervision module only induces a tiny number of computational overheads. The GFLOPs of the Efficient Temporal Transformer module is 6.58 under the settings of $16f$. We recognize that the additional computational overheads of the Efficient Temporal Transformer module mainly come from the temporal transformer encoder sub-module. However, it has been alleviated by the the frame embedding function $h(\cdot)$. The Temporal Sequence Self-supervision module, on the other hand, produces GFLOPs 1.62 with the settings of $16f$. The parameter quantities of the Efficient Temporal Transformer module and the Temporal Sequence Self-supervision modules are very lightweight, 2.20 M and 12.59 M respectively. Furthermore, the module sizes of ETT and TSS are 18.33 MB and 96.12 MB. At the same time, it should be emphasized that the TSS module is only used in the training phase of the model; and in the inference phase, there is only the ETT module. Therefore, the proposed TTSN model is very lightweight and efficient.

\begin{table}[tp!]
    \centering
    \caption{Time complexity and parameter quantity of the Efficient Temporal Transformer module and the Temporal Sequence Self-supervision module of our TTSN.}
    \fontsize{9}{11}\selectfont    
    \begin{tabular}{c|c|c|c}
    \hline
    Module & GFLOPs  &   Params (M) & Size (MB)\\
\hline
    ETT & 6.58 & 2.20 & 18.33 \\
    TSS & 1.62 & 12.59 & 96.12\\
    % \checkmark & \checkmark  & 80.2\\
    \hline
    \end{tabular}\vspace{0.2cm}
    \label{tab:gflops_p}
\end{table}

\noindent\textbf{Single Module Analysis}\,\,To verify the effectiveness of the single module proposed in our TTSN, we have designed and conducted multiple sets of ablation experiments. Respectively, we conduct various ablation experiments on the Efficient Temporal Transformer module and the Temporal Sequence Self-Supervision module in Table \ref{tab:single_module_ana}. We pre-train our TTSN with ImageNet + Kinetics combined datasets and employ the setting of $16f$ with ResNet50 as the backbone. As shown in the Table, when we employ only one of the two modules or both of them simultaneously in our TTSN, the performance of TTSN would improve under all circumstances; particularly, the network performs most desirable when employing both modules in our TTSN.

\begin{table}[tp!]
    \centering
    \caption{Result of the ablation experiment on effectiveness. ``$\checkmark$'' denotes employed, while ``-'' denotes unemployed.}
    \fontsize{9}{11}\selectfont    
    \begin{tabular}{c|c|c}
    \hline
    ETT & TSS  &   Top1.(\%) \\
\hline
    \checkmark & - & 79.3\\
    - & \checkmark & 79.3\\
    \checkmark & \checkmark  & 80.2\\
    \hline
    \end{tabular}
    % \vspace{0.2cm}
    \label{tab:single_module_ana}
\end{table}

\noindent\textbf{Hidden Dimension $l$ Analysis}\,\,We conduct the ablation experiments on the hidden dimension $l$ of temporal transformer encoder sub-module to examine its effects on performance. $l$ is the dimension of 1D frame tensor $\textbf{t}^{j}_{i}$, $j\in\{\alpha,\beta,\gamma\},i\in\{1,2,...,N\}$ and it also represents the hidden dimension of temporal transformer encoder sub-module. According to Figure \ref{fig:3}, we notice the overall performance of the network exhibit an upward tendency with the index of the hidden dimension $l$ keeps on climbing. Our TTSN obtains the best accuracy when the hidden layer dimension $l$ is 12544. The hidden dimension $l$ can not be too small (e.g., $l=49$) to avoid the lack of the effective temporal modeling or too large to avoid the overfitting, so we take $l$=12544 for all experiments in this paper.

\noindent\textbf{Self-Supervision Algorithms Analysis}\,\,To determine the final algorithm for the Temporal Sequence Self-supervision module, we devise four distinct algorithms in total, namely $\mathcal{AA},\mathcal{AR},\mathcal{RA}$ and $\mathcal{RR}$. To examine and analyze the impacts of these algorithms proposed for our TTSN, we conduct a myriad of ablation experiments, reporting ``TOP1. (\%)'' and ``GPU usage'', on the HMDB51 dataset, which is shown in Table \ref{tab:ss-algorithms-ana}. According to Table, we notice that with the gradual increment of randomness, the TOP1. (\%) performance keeps escalating; concurrently, there is a continuous drop in the GPU usage. We assume the \textbf{Randomness level} of the above four algorithms is $\widetilde{R}$, whose relationship is as follows: $\widetilde{R}_{\mathcal{AA}} \textless \widetilde{R}_{\mathcal{RA}} \textless\widetilde{R}_{\mathcal{AR}} \textless \widetilde{R}_{\mathcal{RR}}$. This phenomenon shows that random selection can increase the diversity of samples, thereby further improving the modeling ability of the model in the temporal dimension.

\begin{table}[tp!]
    \centering
    \caption{Result of the ablation experiments of our proposed self-supervision algorithms. The $+$ symbol, for example, $+\mathcal{AA}$ indicates we employ $\mathcal{AA}$ algorithm in the Temporal Sequence Self-supervision module of our TTSN.}
    \fontsize{9}{11}\selectfont   
    \begin{tabular}{c|c|c}
    \hline
    Method & Top1. (\%) & GPU (MB)\\
    \hline
    ETT+$\mathcal{AA}$ & 79.1 & 13967\\
    ETT+$\mathcal{RA}$ & 79.4 & 13967\\
    ETT+$\mathcal{AR}$ & 79.5 & 13675\\
    ETT+$\mathcal{RR}$ & \textbf{80.2} & \textbf{13539}\\
    \hline
    \end{tabular}\vspace{0.2cm}
    \label{tab:ss-algorithms-ana}
\end{table}

\noindent\textbf{Inference Time Analysis}\,\,We report the inference time of our TTSN on RTX 3090, and specify $B=4$, using the setting of $8f$ and $16f$ respectively. The measuring of the inference time takes into account all evaluations, including the consumption of loading data and network inference.
From Table \ref{tab:inference-time-ana}, we can see that whether it is 8 or 16 frames, our TTSN model is able to perform very efficient real-time inference, although it is somewhat slower than some previous models.

\begin{table}[tp!]
    \centering
    \caption{Inference time analysis on a Nvidia RTX 3090 GPU.}
    \fontsize{9}{11}\selectfont    
    \begin{tabular}{c|c|c}
    \hline
    Method & Frames  &Time (ms/video)\\
    \hline
    TSN\cite{2016Temporal} &  8   & 7.9 \\
    TSM\cite{2018TSM}  &  16  &  16.7 \\
    STM\cite{2019STM} &  8  &   11.1 \\
    I3D\cite{2017Quo} &  32  &  2095  \\
    TDN\cite{2020TDN} & 8  &  22.1 \\
    \hline
    \multirow{2}{*}{\textbf{TTSN}}& 8  &  16.1 \cr 
                                  & 16  &  29.38  \\
    \hline
    \end{tabular}
    % \vspace{0.2cm}
    \label{tab:inference-time-ana}
\end{table}

\begin{figure}[tp!]
    \centering
    \includegraphics[width=3.2in]{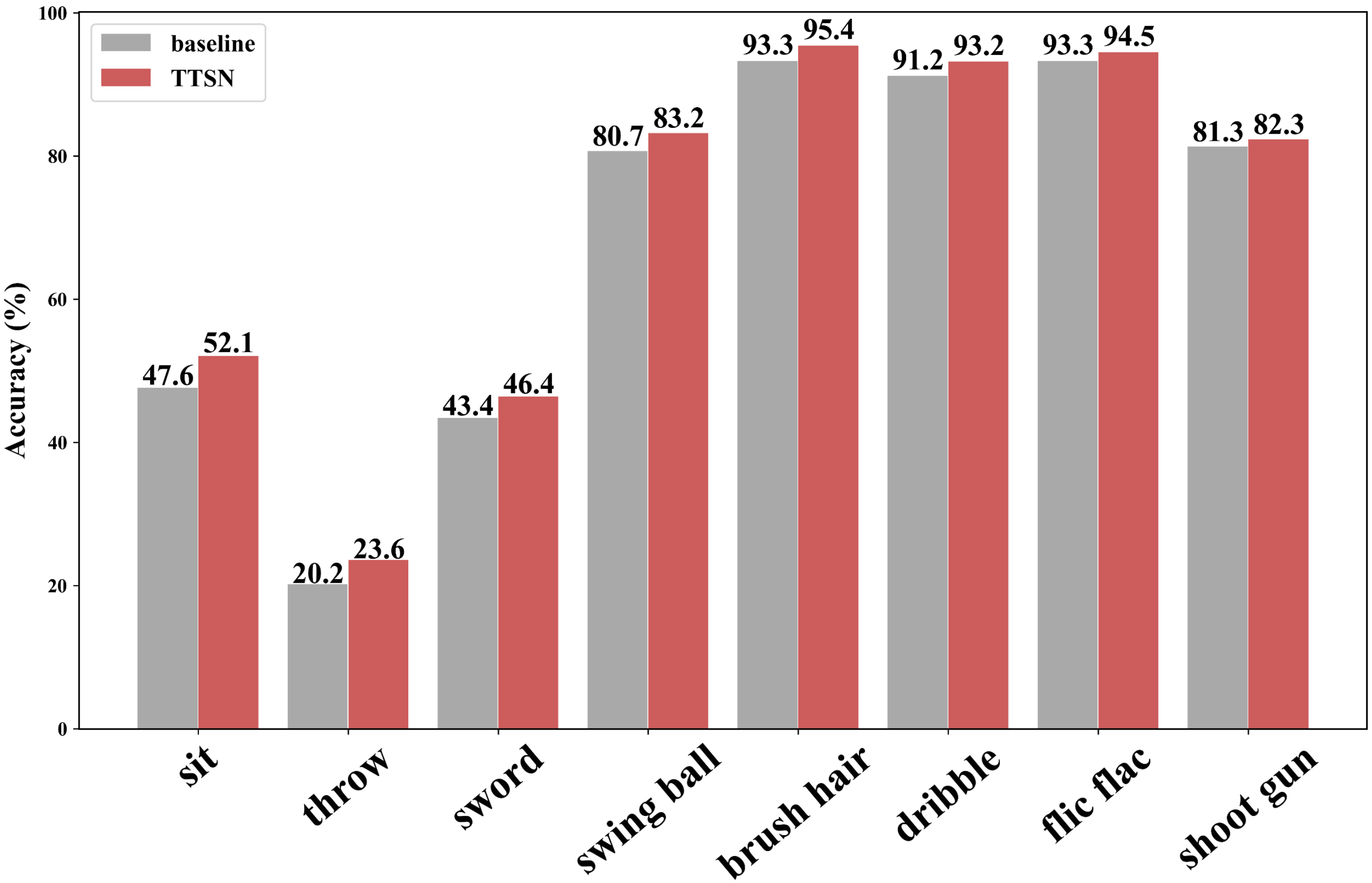}\\
     \caption{Accuracy analysis of some categories on the HMDB51 dataset.}
    \label{fig:accuracy-hmdb51}
\end{figure}
\begin{figure}[tp!]
    \centering
    \includegraphics[width=3.2in]{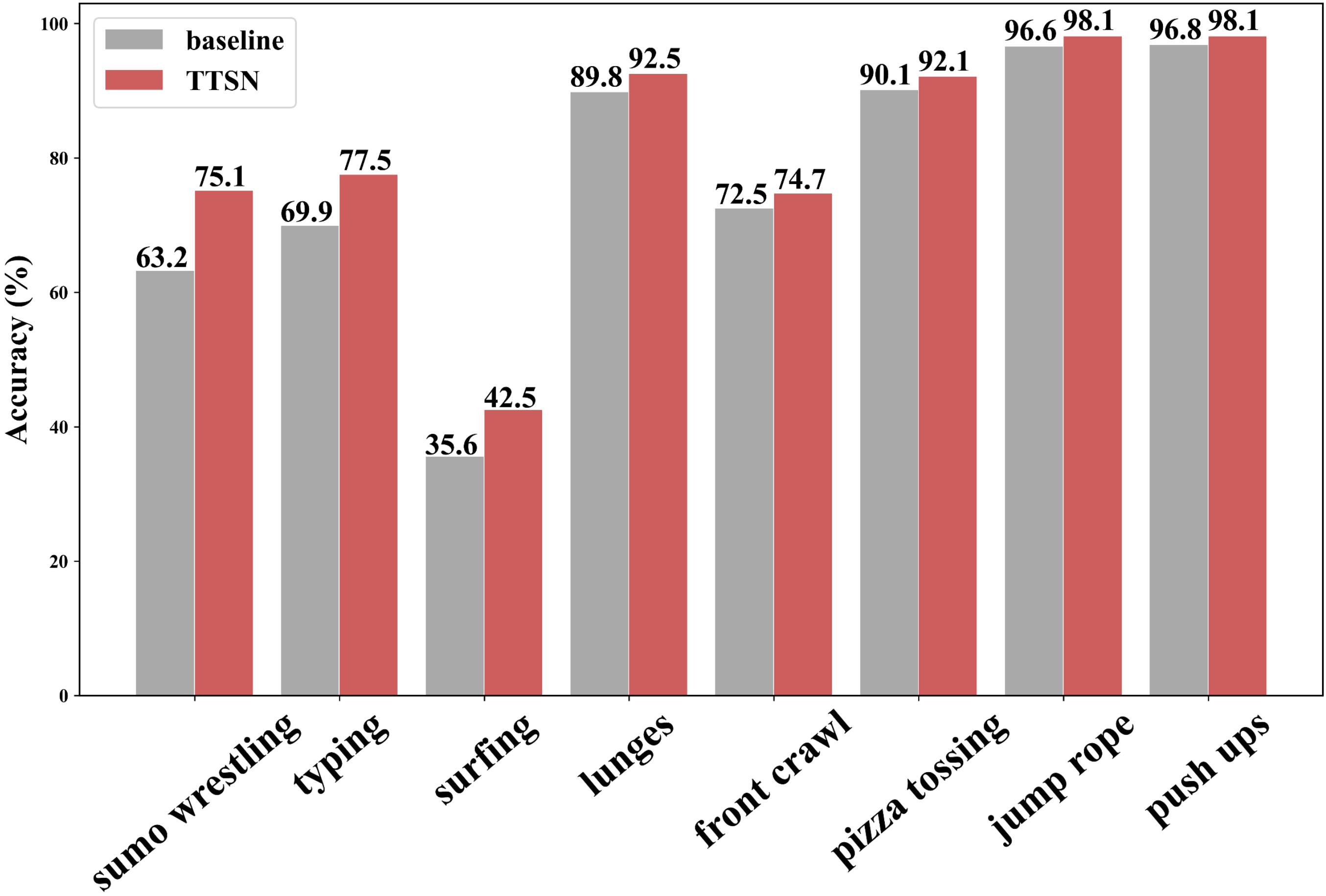}\\
     \caption{Accuracy analysis of some categories on the UCF101 dataset.}
    \label{fig:accuracy-ucf101}
\end{figure}

\noindent\textbf{Accuracy Analysis of Some Categories}\,\,We analyze the accuracy of each category on HMDB51 and UCF101 to find some potential regularities. We would start illustrating by comparing the accuracy of some categories on HMDB51 and UCF101, respectively. From Figure \ref{fig:accuracy-hmdb51} and \ref{fig:accuracy-ucf101}, we notice that the accuracy of some of the confusing actions we have described earlier, such as ``sit'', ``stand'', and ``throw'', has been successfully improved.
We also noticed that the accuracy of distinguishing some complex actions in the time dimension has been improved, such as ``sword'', ``dribble'', ``sumo wrestling'', ``surfing'' and ``push ups''. These phenomena further corroborate our proposed TTSN model, which can enhances the recognition ability of the model for complex and confusing actions by simultaneously modeling complex non-linear relation and inverse motion information in the temporal dimension.

\begin{figure*}[tp!]
\centering
\subfigure[HMDB51]{\includegraphics[height=7cm,width=7.6cm]{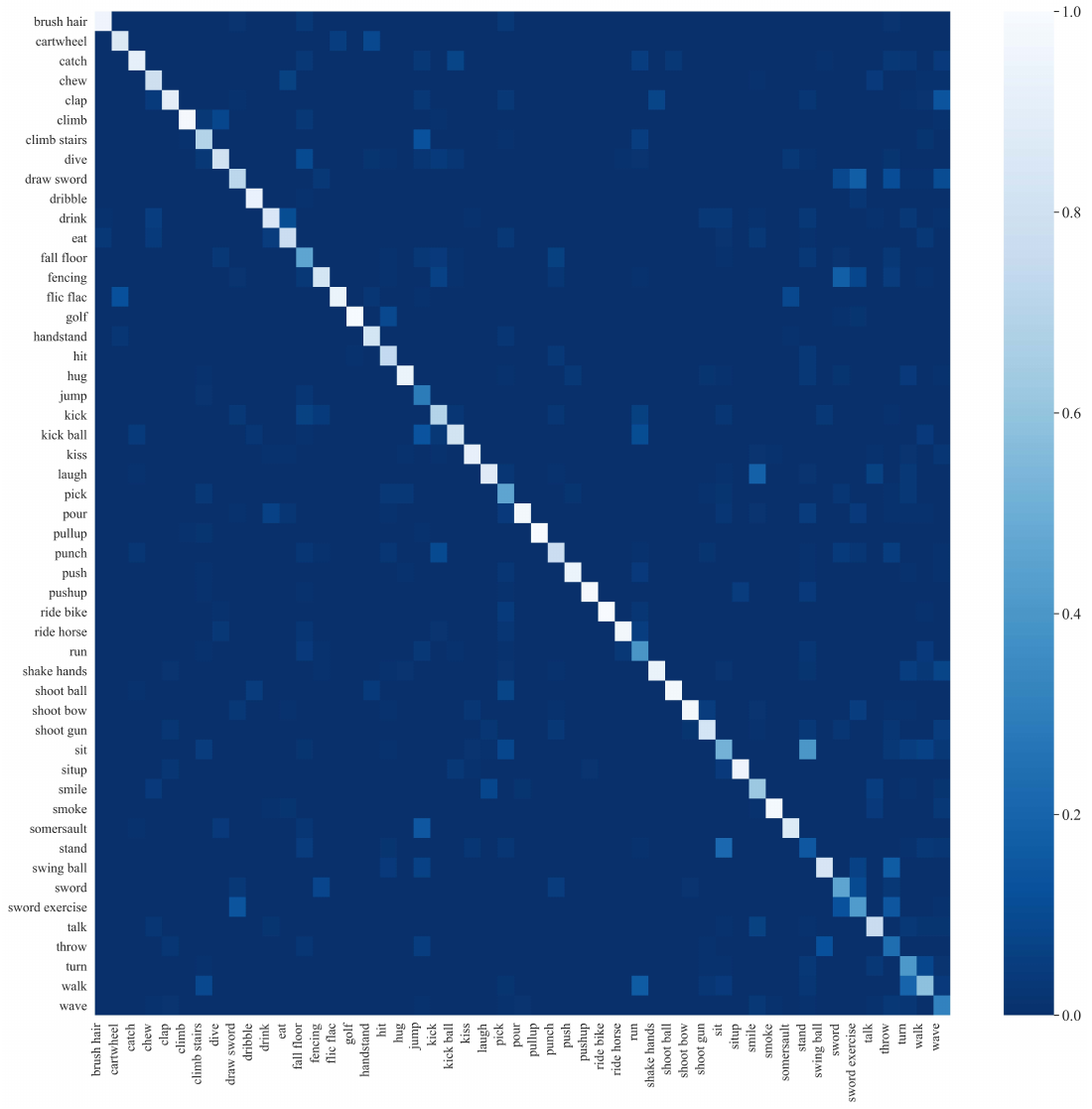}}
\subfigure[UCF101]{\includegraphics[height=7cm,width=7.6cm]{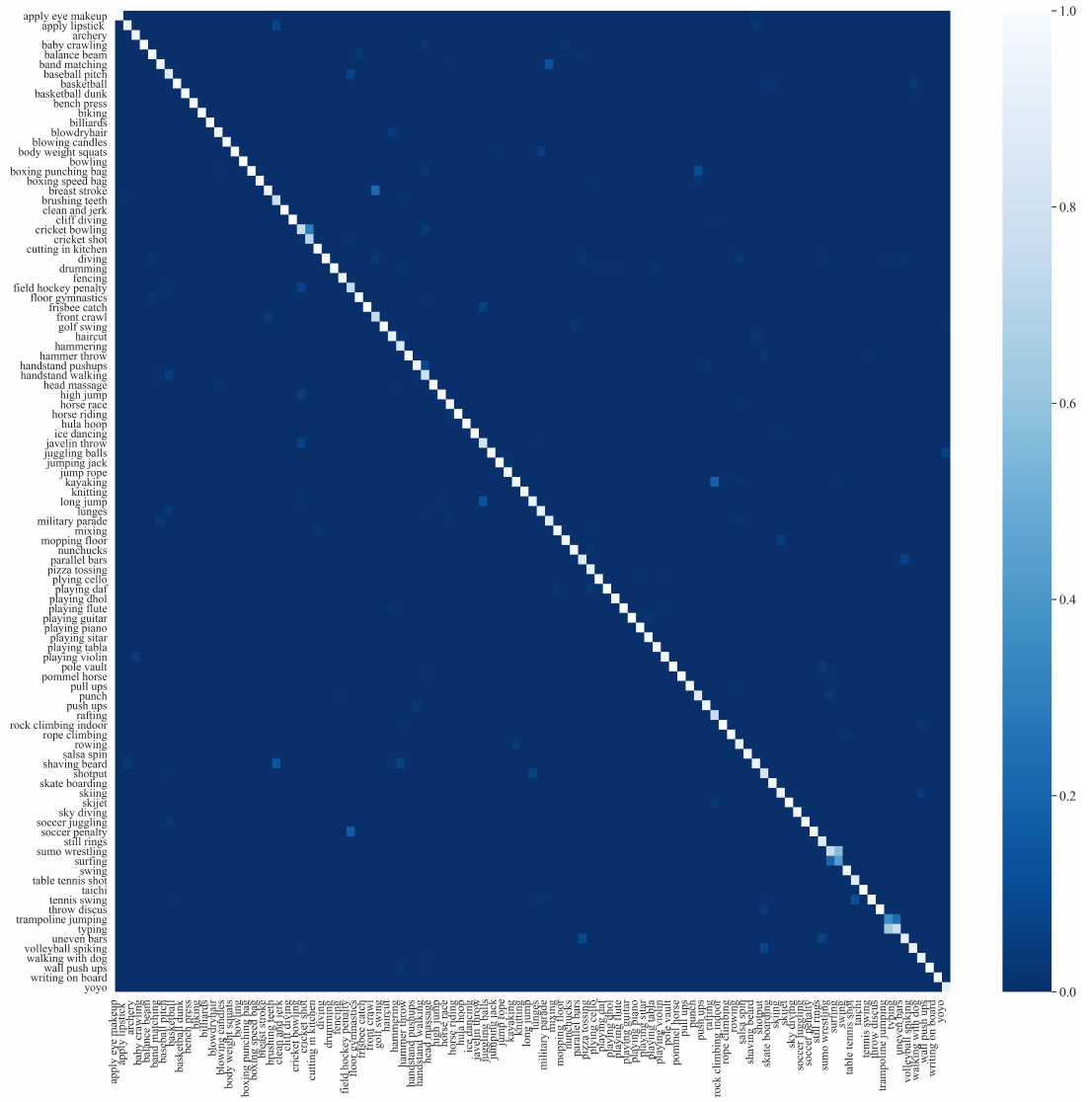}}
\caption{Confusion matrix of our TTSN on HMDB51 and UCF101. Since we take complex non-linear and non-local modeling of the temporal dimension into consideration and introduce the Temporal Sequence Self-supervision module, the confusion matrix of our TTSN outcome is ideal.}
\label{fig:confusion-matrix}
\end{figure*}

\noindent\textbf{Confusion Matrix Analysis}\,\,We present the confusion matrix of our TTSN on HMDB51 and UCF101, respectively. As shown in Figures \ref{fig:confusion-matrix}, our TTSN achieves promising recognition accuracy on both datasets, especially on the UCF101 dataset where it obtains very robust classification accuracy for each category, thanks to the ability of our model to model the temporal relation of the video in a non-linear and bidirectional manner.

\begin{figure*}[tp!]
\centering
\includegraphics[width=\linewidth,height=3.0in]{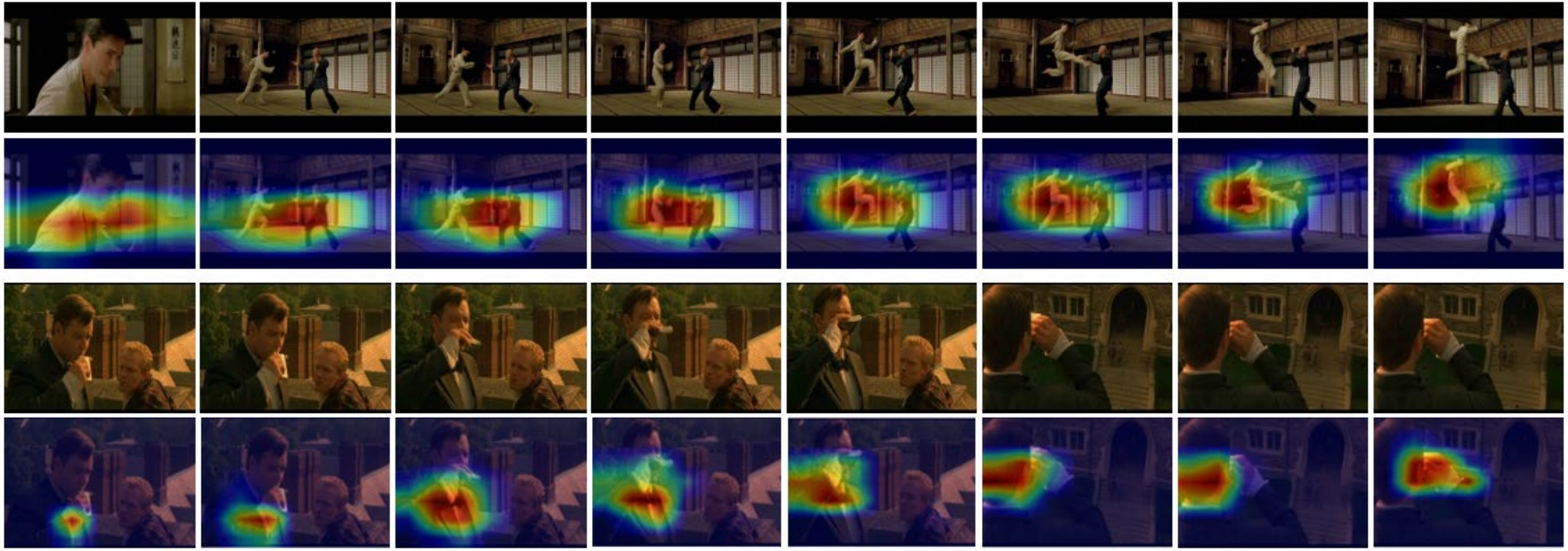}
\caption{The attention maps generated by the Efficient Temporal Transformer Module. We employ $16f$ to visualize on the HMDB51 dataset. \textit{Odd rows: Raw RGB frames. Even rows: Raw RGB frames with our attention maps.}}
\label{fig:vis-ett}
\end{figure*}

\noindent\textbf{Visualization Analysis}\,\,We visualize the obtained attention maps generated by the Efficient Temporal Transformer module, as shown in Figure \ref{fig:vis-ett}. As we expected, the regions that the network pays attention to are those that undergo significant motion changes occur along the temporal dimension, that is, the motion-sensitive pixel regions that we obtained from the non-linear temporal relation modeling.

\section{Conclusion}
In this paper, we introduce TTSN, a novel 2D CNNs-based action recognition network. Our TTSN mainly comprises two modules: an Efficient Temporal Transformer module and a Temporal Sequence Self-Supervision module. The Efficient Temporal Transformer module is employed to efficiently model the non-linear and non-local complex relation of a given temporal dimension and help the network complete the focusing process. The Temporal Sequence Self-Supervision module, on the other hand, constrains the TTSN network to learn robustly and reverse motion information representation from a given temporal dimension. Our TTSN is particularly excellent in learning actions that are inverse to one another in the temporal dimension yet similar in the spatial dimension, such as ``sit'' and ``stand'', or holding a complex non-linear repetitive nature, such as ``brushing hair'' and ``flic-flac''. Principally, by avoiding massive computation of optical flow-based methods and maintaining the complexity of 2D CNNs-based methods, our TTSN is promising because it still contrives to achieve the performance of 3D CNNs-based methods, confirmed by outmatching the state-of-the-art action recognition performance on three mainstream datasets (HMDB51, UCF101, and Something-something V1).

\bibliography{IEEEtran}
\bibliographystyle{IEEEbib}
\end{document}